\pdfoutput=1
\documentclass{article} % For LaTeX2e
\usepackage{iclr2025_conference}
\iclrfinalcopy
\usepackage{amsmath,amsfonts,bm}

\def\eqref#1{equation~\ref{#1}}
\def\1{\bm{1}}

\def\vx{{\bm{x}}}
\def\vy{{\bm{y}}}

\def\mA{{\bm{A}}}
\def\mB{{\bm{B}}}

\def\mI{{\bm{I}}}

\def\mV{{\bm{V}}}
\def\mW{{\bm{W}}}
\def\mX{{\bm{X}}}
\def\mY{{\bm{Y}}}
\def\mZ{{\bm{Z}}}

\DeclareMathAlphabet{\mathsfit}{\encodingdefault}{\sfdefault}{m}{sl}
\SetMathAlphabet{\mathsfit}{bold}{\encodingdefault}{\sfdefault}{bx}{n}

\newcommand{\R}{\mathbb{R}}

\usepackage{amssymb}
\usepackage{hyperref}
\usepackage{url}
\usepackage{booktabs}
\usepackage{multirow}
\usepackage{graphicx}
\usepackage{amsthm}
\usepackage{mdframed}
\usepackage{cleveref}
\usepackage{listings}

\usepackage{floatrow}
\newfloatcommand{capbtabbox}{table}[][\FBwidth]
\theoremstyle{definition}
\newtheorem{definition}{Definition}
\newtheorem{proposition}{Proposition}
\crefname{lstlisting}{listing}{listings}
\Crefname{lstlisting}{Listing}{Listings}
\title{ALLoRA: Adaptive Learning Rate Mitigates LoRA Fatal Flaws}

\author{Hai Huang \\
Google \\
1600 Amphitheatre Parkway \\
Mountain View, California \\
\texttt{haih@google.com} \\
\And
Randall Balestriero \\
Department of Computer Science \\
Brown University \\
\texttt{rbalestr@brown.edu}
}

\begin{document}

\maketitle

\begin{abstract}
Low-Rank Adaptation (LoRA) is the bread and butter of Large Language Model (LLM) finetuning. LoRA learns an additive low-rank perturbation, $\mA\mB$, of a pretrained matrix parameter $\mW$ to align the model to a new task or dataset with $\mW+\mA\mB$. We identify three core limitations to LoRA for finetuning--a setting that employs limited amount of data and training steps. First, LoRA employs Dropout to prevent overfitting. We prove that Dropout is only suitable for long training episodes but fails to converge to a reliable regularizer for short training episodes. Second, LoRA’s initialization of $\mB$ at $0$ creates a slow training dynamic between $\mA$ and $\mB$. That dynamic is also exacerbated by Dropout that further slows the escape from $0$ for $\mB$ which is particularly harmful for short training episodes. Third, the scaling factor multiplying each LoRA additive perturbation creates ``short-sighted'' interactions between the LoRA modules of different layers. Motivated by principled analysis of those limitations, we find an elegant solution: a Dropout-free, scaling-free, LoRA with Adaptive Learning rate--coined ALLoRA. By scaling the per sample and per parameter gradients with a coefficient inversely proportional to parameters’ $\ell_2$ norm, ALLoRA alleviates those three limitations. As a by-product, ALLoRA removes two hyper-parameters from LoRA: the scaling factor and the dropout rate. Empirical results show that ALLoRA admits better accuracy than LoRA on various settings, including against recent LoRA variants such as Weight-Decomposed Low-Rank Adaptation (DoRA). Ablation studies show our solution is the optimal in a family of weight-dependent / output-dependent approaches on various LLMs including the latest Llama3.
\end{abstract}

\section{Introduction}

Large Language Models (LLMs) \citep{hoffmann2022training, touvron2023llama, jiang2023mistral} are Deep Neural Networks (DNNs)--commonly built from Transformer with self-attention--built for sequence processing, e.g., Natural Language Processing (NLP). LLMs have radically changed the way we approach NLP \citep{chowdhary2020natural} by removing the need for handcrafted feature engineering such as bags of words \citep{zhang2010understanding}. Instead, current solutions directly operate on the input data--or a lossless compression known as tokens \citep{shibata1999byte}. Because we now have access to humongous amount of text data, the standard training pipeline for LLMs take the following form. First, the LLM is {\em pretrained} on a large text corpus through next-token prediction. That autoregressive pretext-task enables the LLM to learn the underlying dynamic of the language. Commonly, RLHF is also employed after pretraining to make the model's behavior shift from autoregressive to agentic. Then, the LLM is {\em fine-tuned} on a more specific downstream task or dataset. That fine-tuning is user-specific and plays a fundamental role in making LLMs practically useful but relies on much more limited datasets, as we formalize below.

\begin{mdframed} {\em \textbf{Premise:} The training regime involved in pretraining and fine-tuning are drastically different. The former employs limitless data and abundant training steps while the latter employs limited data and few training steps.} \end{mdframed}

That premise is now widely accepted upon as the latest state-of-the-art LLM solutions stem from the numerous open-source industry groups such as Meta's Llama \citep{dubey2024llamaf}, Google's Gemma \citep{team2024gemma}, Apple's OpenELM \citep{mehtaOpenELMEfficientLanguage2024}, or Cohere's Aya. Hence, as LLM practitioners, most of the attention is now turning into deriving fine-tuning strategies that meet the very particular needs of fine-tuning LLMs.

To tackle that paradigm shift introduced by the pretraining-finetuning strategy, specialized methods have been developed, such as the eponymous Low-Rank Adaption (LoRA). LoRA has fueled countless deployment of LLMs--as it took a gigantic leap in accommodating for the fine-tuning regime. In short, LoRA proposes to fine-tune a LLM by learning an additive low-rank matrix perturbation to some of the LLM's internal parameter matrices. Core to its design, LoRA leverages (i) Dropout \citep{srivastava2014dropout} as a mean to prevent overfitting to the fine-tuning task, and (ii) zero-initialization to ensure that training starts from the LLM's pretrained mapping, and (iii) a scaling factor that rescales the LoRA's matrix factorization. While the impact of LoRA is ubiquitous, we nonetheless believe that LoRA could be further improved based on three observations.

\begin{mdframed} {\em \textbf{LoRA's three fatal flaws for finetuning:} First, \textbf{Dropout}--a stochastic regularizer--whose benefits quickly vanish when considering fine-tuning, and can in fact introduce detrimental additional variance to the training. Second, the \textbf{zero-initialization} which is difficult to escape from due to Dropout's implicit regularization. Third, the \textbf{scaling parameter} that introduces nonlinear interactions between LoRA modules of different layers.} 
\end{mdframed}

While each of those three design choices are well-motivated when considering long training, e.g., pretraining, it becomes harder to prove their benefits when considering fine-tuning that only employs a minimal amount of training steps. That is why, after carefully bringing to light and studying the above flaws of LoRA--in the context of fine-tuning--in \cref{sec:flaw}, we will propose a novel variation of LoRA that we coin \textbf{ALLoRA} for \textbf{A}daptive \textbf{L}earning rate \textbf{LoRA} (\cref{sec:allora}). ALLoRA proposes to remove the Dropout regularizer and the scaling factor while adding an adaptive learning rate for the low-rank matrices entries. As depicted in \cref{code:allora}, the implementation is straightforward with theoretical and practical benefits. First, by removing the Dropout regularization and the scaling factor, ALLoRA is simpler to employ as it no longer requires cross-validation of those parameters. Second, we demonstrate how ALLoRA improves performances over LoRA and alternatives such as DoRA. In short, our adaptive learning rate strategy is able to prevent over-fitting, learn competitive solutions, and converge more quickly than alternatives--all while employing less hyper-parameters.

% Problems:
% \begin{itemize}
%     \item A key difference between finetuning
% and pretraining is that finetunining is limited by a small number of epochs over a small finetuning dataset. Running more epochs risks overfitting the small dataset. As revealed by \cite{liu2024dora}, it is such a waste if $|| BAx ||$ could not make any effective
% impact over $||W^\ast x||$ during a good portion of the limited process. hence scaling factor
%     \item Ironically, dropout (\cite{srivastava2014dropout}) is still applied to LoRA to combat overfitting, albeit at a much lower dropout rate. Applying dropout means that a constant fraction of weights will never receive updates during training, yet another waste as the number of parameters is already highly confined.
%     \item Last but not least, having too many hyperparameters: learning rate, scaling factor, and dropout rate, further complicates the
%     finetuning process.
% \end{itemize}

We summarize our contributions below:
\begin{enumerate}
  \item We identify three inefficiencies (\cref{sec:dropout,sec:landscape,sec:scaling}) in the current LoRA design making it unfit for short training, i.e., finetuning, that we empirically validate in \cref{sec:validation}.
  \item We propose a novel adaptive learning rate variation of LoRA--coined ALLoRA--free of two of the original LoRA's complicated designs: the Dropout regularizer and the scaling factor (\cref{sec:allora}).
  \item We empirically validate the benefits of ALLoRA over numerous datasets and model architectures including the latest Llama3 family (\cref{sec:percept,sec:commonsense}). We obtain that despite ALLoRA employing less hyperparameters than LoRA, it is able to outperform its counter part and recent variants such as DoRA consistently.
\end{enumerate}

The full codebase to reproduce figures and tables is publicly available\footnote{\url{http://github.com/rbalestr-lab/allora}} along side the finetuned models' weights\footnote{\url{http://bit.ly/allora-weights}}.

% Other adaptive functions.

% Mention efficiency ~= accuracy, given the restriction on limited \# finetuning steps, which is often \# epochs * size of dataset.

% Mention adaptive learning rate may have applications beyond LoRA.

% Mention generalized framework may have applications beyond LoRA.

\section{Related Works}
LoRA is a type of \textit{Parameter Efficient Fine Tuning} (PEFT) designed to reduce the cost of finetuning, especially with LLMs. As LLMs typically have large number of parameters--in the scale of billions--one can not afford to finetune all those parameters on a particular downstream task or dataset. Existing PEFT can be divided into three categories, namely \textit{Adapter-based Methods}, \textit{Prompt-based Methods}, and \textit{LoRA}.

Adapter-based methods \citep{houlsby2019parameter, he2022towards, karimi2021hyper, karimi2021parameterefficient} introduce additional trainable modules, \textit{a.k.a.} the \textit{adapters}, into the original backbone whose weights are frozen during the finetuning. In \cite{houlsby2019parameter}, linear modules were added in sequence to the existing layer, while in \cite{he2022towards}, they were added in parallel to the existing layer for the sake of better performance. 

Prompt-based methods \citep{lester2021emnlp, razdaibiedina2023residual, wang2023non} introduce soft tokens as trainable parameters and prepend them to the prompt. This category is the least intrusive as the finetuning can be done by only prompting the LLMs. 
However, prompt-based methods are in general sensitive to initialization and their overall effectiveness is affected. 

LoRA \citep{hu2021lora} uses low-rank matrices to simulate weight changes of the pretrained weights. Since  low-rank matrices can be merged back to original weights, LoRA does not incur any additional cost at inference, which is a significant advantage over the other two categories. Many variants were proposed lately. For example, in \cite{zhang2023adaptive}, SVD decomposition was employed to determine 
significance of singular values, and less important ones are pruned. \cite{hyeonwoo2022lowrank} applies low-rank Hadamard product to federated learning. \cite{qiu2023control} and \cite{liu2024param} adopt orthogonal factorization and applied to diffusion models. \cite{renduchintala2023tied} introduces weight tying and realizes more savings on number of parameters. A unified LoRA family was introduced for Stable diffusion in \cite{yeh2024text}. Different combinations of LoRA are chosen for different tasks in \cite{ponti2022combining}. A scaling vectors is learnt to adjust a pair of frozen random matrices shared across layers in \cite{kopiczko2024vera}. 

More recently, \cite{liu2024dora} proposes decomposing the weights into directional and magnitude components to boost accuracy. 
\cite{hayou2024impact} studies the optimal initialization of the low-rank matrices, and a follow-up work \citep{hayou2024lora+} proposes to apply different learning rate to different low-rank matrices. Superficially this is similar to one of our idea to adapt learning rate, though our idea is inspired by a principled study of dropout \citep{srivastava2014dropout}.

More broadly, \cite{zhao2024galore} applies the low-rank concept to compute low-rank gradients directly. \cite{jang2024lora} provides a study on the existence and convergence of LoRA solutions. And \cite{zhang2024riemannian} is a study of the potential ill conditioned low-rank matrices.

\section{A Critical Analysis Of LoRA for Finetuning}
\label{sec:flaw}

PEFT is the main bridge between large pretrained LLMs and specialized practical use-cases. Hence, PEFT research is extremely active which led to numerous variations of LoRA being developed. Our theoretical study builds on the seminal version formalized below, other LoRA variants will be compared in our experimental \cref{sec:percept,sec:commonsense}.

\begin{definition} (Low Rank Adapters (LoRA) from \cite{hu2021lora}). For any weight matrix $\mW \in \R^{n_1 \times n_2}$
in the pretrained model, we constrain its update in the finetuning process by representing the latter with a low-rank
decomposition $\tilde{\mW} = \mW + \frac{\alpha}{r}\mB\mA$. Here, only the weight matrices $\mB \in \R^{n_1\times r}$, $\mA \in \R^{r\times n_2}$
are trainable. The rank $r \ll \min(n_1, n_2)$ and $\alpha \in \R$ are tunable constants.
\label{def:lora}
\end{definition}

Despite LoRA's simplicity, very little attention was put on its core premises within a finetuning context, i.e., with limited amount of training steps. We propose here a principled and critical study of LoRA, culminating with our finding of three core flaws of LoRA for short training (\cref{sec:dropout,sec:landscape,sec:scaling}). The following \cref{sec:allora} will investigate our solution, ALLoRA.

\subsection{First Flaw: A Stochastic Regularization that Will not Converge}
\label{sec:dropout}

LoRA's regularization comes from Dropout \citep{srivastava2014dropout}, i.e., applying a multiplicative random binary mask at the bottleneck of the matrix factorization. The use of Dropout would seem justified since \cite{wager2013dropout} showed in the linear regime that Dropout acts as a variant of $\ell_2$ regularization on normalized design matrix (inputs), a result also found in the previous study of \cite{wang2013fast}. However, those theoretical results deal with {\em expectations}, i.e., infinite training time. Similarly, previous empirical studies of Dropout showed great regularization benefits, but all those studies only considered full training, i.e., long training episodes that ultimately converge to their expectation. We argue that those beneficial findings {\em do not hold during fine-tuning which only employs limited training steps}.

To understand the impact of Dropout in terms of regularization, we will consider a few variations of models and study the discrepancy between the {\em expected} benefit of Dropout with its {\em empirical} realisations. We will conduct validation on real dataset and LLMs to support our theory throughout the section, and in particular in \cref{sec:validation}. Also, we consider here and in \cref{sec:landscape} $\alpha/r$ to be $1$ without loss of generality, none of our results are impacted by that constant scaling factor.

{\bf Linear model with full training.}~Let's first consider the original setting of a linear model with Dropout applied to its predictions. That is, we consider the following Ordinary Least Squares (OLS) setting $\| \mY - (\mX\mW) \odot \mV\|_F^2$, 
with $\mY \in \mathbb{R}^{N\times C},\mX \in \mathbb{R}^{N\times D},\mW \in \mathbb{R}^{D\times C}$ and the random realization of dropout matrix $\mV \in \{0,\frac{1}{p}\}^{N \times C}$. We note that such parametrization of Dropout is commonly employed in the literature to ensure that its expectation is equal to $1$, and it is also PyTorch's official implementation. We have the following property that has motivated the use of Dropout through more than a decade by now:
\begin{align*}
    \mathbb{E}&\left[\| \mY - (\mX\mW) \odot \mV\|_F^2\right]\\
    &=
     \|\mY\|_F^2+\mathbb{E}\left[-2Tr\left(\mY \left(\mV^\top\odot(\mX\mW)^\top\right)\right) + Tr\left(\left(\mV^\top\odot(\mX\mW)^\top\right)\left((\mX\mW) \odot \mV\right)\right) \right]\\
    &=
    {\color{blue}\|\mY\|_F^2-2Tr\left(\mY (\mX\mW)^\top\right) + Tr\left((\mX\mW)^\top(\mX\mW) \right)}+{\color{orange}\left(\frac{1}{\lambda}-1\right)Tr\left((\mX\mW)^\top(\mX\mW) \right)},
\end{align*}
which is solved for $\mW = p(\mX^\top\mX)^{-1}\mY^\top\mX$. We note that we decomposed the loss into the terms that happen in the original (Dropout-free) setting, in {\color{blue} blue}, and the Dropout induces terms in {\color{orange} orange}. Comparing that with the usual Tikhonov regularization that produces $\mW = (\mX^\top\mX+\lambda \mI)^{-1}\mY^\top\mX$ we see then whenever the eigenvalues of $\mX^\top\mX$ are all identical to a positive constant $c$ (e.g. when $\mX$ is whitened), $\mW = \frac{c}{c+\lambda}(\mX^\top\mX)^{-1}\mY^\top\mX$ hence recovering the dropout solution. Assuming $c=1$ without loss of generality, we obtain that applying Dropout with rate $p=\frac{1}{1+\lambda}$ is equivalent to applying Tikhonov regularization with rate $\lambda$. As a result, we see that if training for long enough, Dropout can efficiently replace explicit forms of regularization such as weight decay. While the above results recovers known theoretical analysis of Dropout--showing its benefits as an implicit regularizer--those derivations only emerge from taking expectation of the loss, i.e., considering infinite training steps.
For us, the question thus turns into the following: {\em what are the benefits of Dropout as a LoRA regularizer for very short training regime such as finetuning?}

\begin{figure}[t!]
    \centering
    \includegraphics[width=\linewidth]{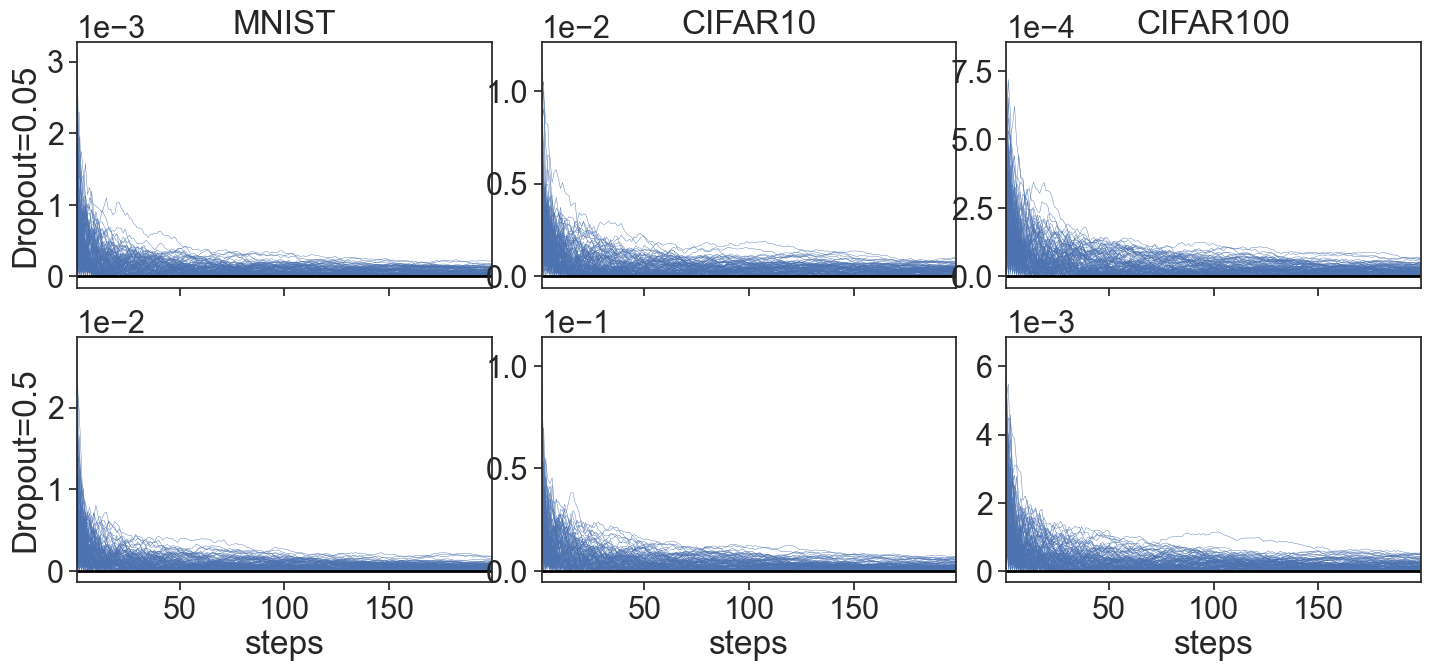}
    \caption{\small We depict the absolute difference ({\bf y-axis}) between the empirical and expected finetuning LoRA loss with varying Dropout rates ({\bf rows}) on different datasets ({\bf columns}) as a function of the number of Dropout realisation ({\bf x-axis}). We observe that regardless of the dataset and Dropout probability, the empirical error is a poor estimate of the true expected loss even after hundreds of averaged realisations. Hence, {\bf finetuning with Dropout produces a large amount of random noise that go well beyond its regularization benefit which only emerges after a large number of steps}. That finding is also confirmed by the LLM experiment in \cref{fig:accuracy-epoch-dropout}.}
    \label{fig:convergence_lora}
\end{figure}

{\bf LoRA with Dropout fails to converge.}~Let's denote the LoRA linear finetuning setting as follows
\begin{align}
\mathcal{L} \triangleq \| \mY - \mX LoRA_{\mA,\mB}(\mW)\|_F^2=\| \mY - \mX\mW - ((\mX\mA)\odot \mV)\mB\|_F^2.
\end{align}
In the above setting, we can slightly modify the above analysis to obtain the following result
\begin{align*}
    \mathbb{E}&\left[\mathcal{L}\right]\hspace{-0.1cm}=\hspace{-0.1cm}{\color{blue}\|\mY \hspace{-0.1cm}-\hspace{-0.1cm} \mX\mW\|_F^2-2Tr\left( (\mY \hspace{-0.1cm}-\hspace{-0.1cm} \mX\mW)^\top\hspace{-0.1cm}(\mX\mA\mB)\right) \hspace{-0.1cm}+ \hspace{-0.1cm}\| \mX\mA\mB\|_F^2}{\color{orange}+\frac{1-\lambda}{\lambda}\sum_{r=1}^{R}\left\| \mX (\mA)_{.,r} (\mB)_{r,.}\right\|_F^2},
\end{align*}
where we recall that $\frac{1}{\lambda} -1 > 0, \forall \lambda \in (0,1)$. As for the full training setting, we decomposed the loss into the terms that happen in the original (Dropout-free) setting, in {\color{blue} blue}, and the Dropout induces terms in {\color{orange} orange}. Before diving into the impact of Dropout in terms of regularization, it is interesting to understand how many samples or training steps are needed for the empirical estimate to converge to the above expectation. We propose that analysis in \cref{fig:convergence_lora}. We clearly see that even for small Dropout value $0.05$, the convergence is quite slow on the various settings we explored.

\begin{figure}[t!]
    \centering
    \begin{minipage}{0.49\linewidth}
    \includegraphics[width=1\linewidth]{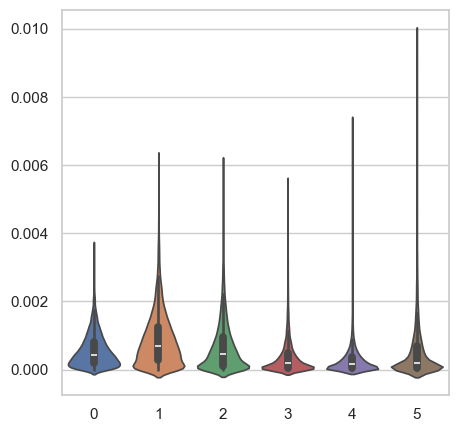}
    \end{minipage}
    \begin{minipage}{0.49\linewidth}
    \includegraphics[width=1\linewidth]{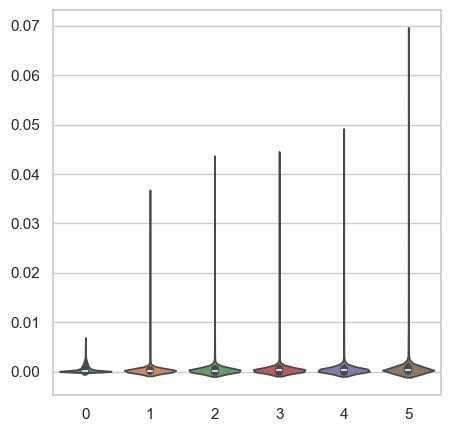}
    \end{minipage}
    \caption{\small Depiction of the distribution of standard deviation of gradients ({\bf y-axis}) w.r.t. the second layer of a MLP trained for MNIST ({\bf left}) and CIFAR10 ({\bf right}) classification, equipped with Dropout. At each training epoch ({\bf x-axis}), we consider a single mini-batch and compute the gradients under numerous Dropout realisation. For each entry in the matrix of gradients, we compute the standard deviation and report the distribution over entries. We clearly see that {\bf while the average variance of the gradient decreases slightly during training, the tail significantly increases, leading to unstable training in finetuning regimes}.}
    \label{fig:lora_dropout}
\end{figure}

To better understand those trends, we propose a first simple bound on ``how far off'' is the expectation to the empirical estimator as a function of the number of realisations, or, steps. The detailed derivations as provided in \cref{proof:upper_bound}.We obtain
\begin{multline*}
    \mathbb{E}\left[ \left| \frac{1}{N}\sum_{n=1}^{N}\| \mY - \mX\mW - ((\mX\mA)\odot \mV_n)\mB\|_F^2-\mathbb{E}\left[\| \mY - \mX\mW - ((\mX\mA)\odot \mV)\mB\|_F^2\right]\right|\right]\\
    \leq\frac{Std\left[\| \mY - \mX\mW - ((\mX\mA)\odot \mV)\mB\|_F^2\right]}{\sqrt{N}}.
\end{multline*}
As a result, the benefit of $N$, in our case the number of training steps, only appears for large $N$ as depicted in \cref{fig:convergence_lora}. This is particularly true as training progresses where the value of the standard deviation of the error ($Std\left[\| \mY - (\mX\mW) \odot \mV\|_F^2\right]$) may increase under the Dropout. While this could be dataset and model specific, we illustrate the distribution of that random variable during training in \cref{fig:lora_dropout}. We look at the standard deviation of the gradients as a function of Dropout realisation in a simple MLP with MNIST classification task as training progresses and observe that the distribution becomes more and more heavy-tailed as training progresses, indicating that some mini-batch may receive highly noisy loss and gradient updates, hence making it hard to recover during short finetuning settings.

\subsection{Second Flaw: Zero Initialization and Unfair Regularization}
\label{sec:landscape}

The second flaw we uncover comes from the zero initialization of $\mB$. As we will see, that is a limitation regardless of employing Dropout or not, but Dropout exacerbates it. 

{\bf Zero initialization implies imbalanced training dynamics.}~A peculiarity of LoRA compared to most other deep learning framework lies in its asymmetric initialization. While $\mA$ is initialized with random entries, $\mB$ is initialized at $0$. This initialization is intuitive when looking at the output of LoRA being $0$ at first, i.e., one starts from the original model and then moves away from that if needed for the finetuning task. However, this seemingly reasonable initialization creates a strong imbalance in the training dynamic of $\mA$ and $\mB$. To see that, we can extend our derivation to obtain the derivative of the expected loss as
\begin{align*}
    \nabla_{\mA} \mathbb{E}\left[\mathcal{L}\right]=& -\mX^\top \left(\mY-\mX\mW\right)\mB^\top+\mX^\top\mX \mA  \left(\frac{1-\lambda}{\lambda}\text{diag}(\| \mB_{1,:}\|_2^2,\dots,\| \mB_{r,:}\|_2^2)+\mB\mB^\top\right),\\
    \nabla_{\mB} \mathbb{E}\left[\mathcal{L}\right]=& -\mA^\top\mX^\top \left(\mY-\mX\mW\right)\\
    &+\left(\mA^\top\mX^\top\mX \mA+\frac{1-\lambda}{\lambda}\text{diag}(\| (\mX\mA)_{1,:}\|_2^2,\dots,\| (\mX\mA)_{r,:}\|_2^2)\right)\mB,
\end{align*}
hence the effective gradient norm for $\mA$ is $0$ during the first few steps because of $\mB$ being $0$, while the gradient norm for $\mB$ will be high $\|\mA^\top\mX^\top \left(\mY-\mX\mW\right) \|_F^2 \gg 0$--regardless of the Dropout rate employed. While such issue only concerns the first few training steps, it is clear that it will be detrimental in a finetuning regime where the total number of training steps is limited. We note that if $\mA$ is zero-initialized instead of $\mB$, our entire argument still holds as the same slow training dynamic appears albeit with respect to $\mA$ instead of $\mB$. Our findings support the conclusion of \cite{hayou2024lora+} that demonstrated how $\mA$ and $\mB$ should receive different learning rates to improve LoRA's performances.

\begin{figure}[t!]
    \centering
    \includegraphics[width=\linewidth]{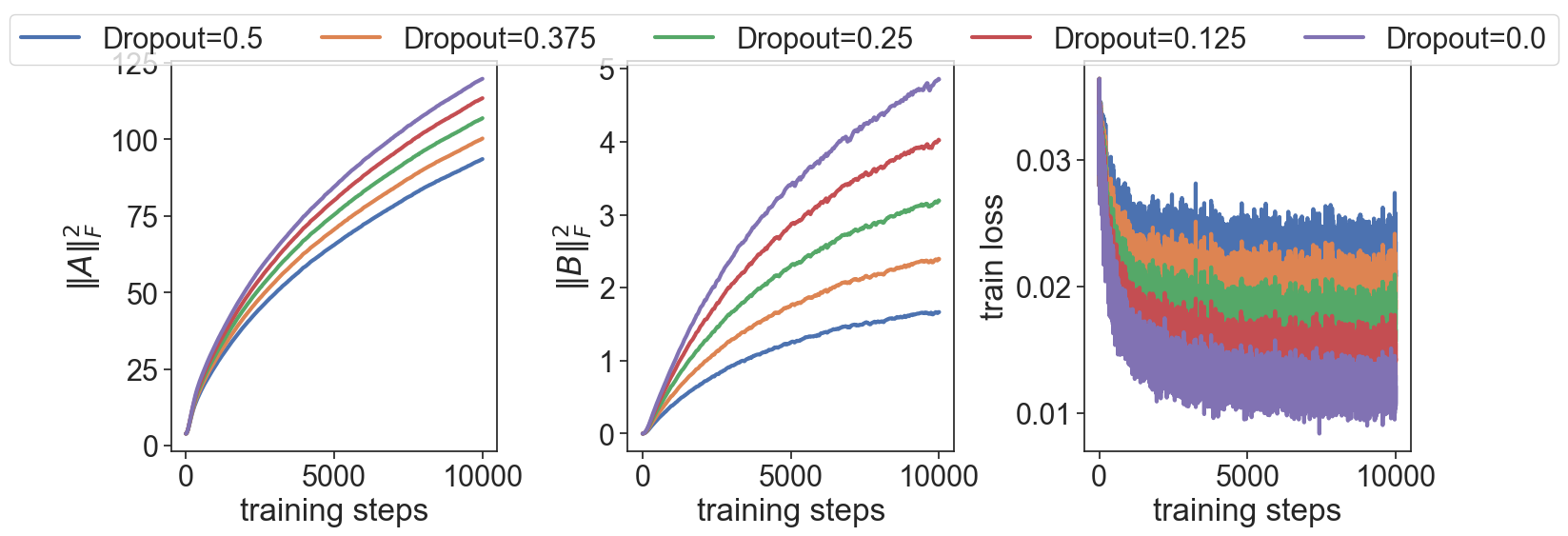}
    \caption{\small Depiction of the norm of $\mA$ ({\bf left}), the norm of $\mB$ ({\bf middle}) and training loss ({\bf right}) for a MNIST LoRA finetuning experiment. We see that as training progresses ({\bf x-axis}) as the impact of increased Dropout probability ({\bf colors}) has a {\bf disproportionate regularization impact on $\mB$ while barely impacting the norm of $\mA$, indicating an asymmetry in Dropout's implicit regularization that makes LoRA slow to train}.}
    \label{fig:lora_norms}
\end{figure}

{\bf Dropout further slows down the escape from $0$.}~An additional issue arises when using Dropout. In that setting, the escape of $\mB$ from $0$, which is needed for $\mA$ to also learn and for the LoRA module to be effective, will be further slowed down. To better characterize that effect, let's study the close-form regularization impact of Dropout using the expected loss derived in \cref{sec:dropout}. 
As per the full training regime, we see that Dropout acts as a regularization on $\mA$ and $\mB$ based on their alignment with $\mX$. In fact, taking derivative with respect to the loss, we see that the orange contributing term's gradient is given by 
\begin{align*}
    \nabla_{\mA}\left(\frac{1}{\lambda}-1\right)\sum_{i=1}^{r}\left\| \mX (\mA)_{:,i} (\mB)_{i,:}\right\|_F^2 &= \left(\frac{1}{\lambda}-1\right)\mX^\top\mX \mA  \text{diag}(\| \mB_{1,:}\|_2^2,\dots,\| \mB_{r,:}\|_2^2),\\
    \nabla_{\mB}\left(\frac{1}{\lambda}-1\right)\sum_{i=1}^{r}\left\| \mX (\mA)_{:,i} (\mB)_{i,:}\right\|_F^2 &= \left(\frac{1}{\lambda}-1\right)\text{diag}(\| (\mX\mA)_{1,:}\|_2^2,\dots,\| (\mX\mA)_{r,:}\|_2^2)\mB.
\end{align*}
Hence, Dropout with LoRA regularizes the matrices $\mA$ and $\mB$ using a weight decay type of regularization but weighted by the squared $\ell_2$ norm of the rows of $\mB$ and the squared $\ell_2$ norm of the columns of $\mX\mA$, respectively. Those findings bring to light an unfair regularization impact of Dropout with LoRA. At initialization, i.e., when $\mB=0$, the strength of the regularizer on $\mA$ will be null. Yet, the strength of the regularizer on $\mB$ is large as it is equal to the norm of $\mX\mA$ which, even at initialization, will be largely greater than $0$. This is concerning since not only $\mB$ starts from a zero-norm initialization opposed to $\mA$, but also its regularization is stronger than that of $\mA$. We depict that dynamic in a practical scenario in \cref{fig:lora_norms}. We observe that the impact of Dropout's regularization, as measured by varying the value of $\lambda$, is minimal on the training dynamic of $\mA$, only altering the final norm by about 30\%, while the impact of Dropout's regularization on $\mB$ is drastic, producing solutions with norms varying by more than 500\%.

\subsection{Third Flaw: Ripple Effect Of Scaling Factor}
\label{sec:scaling}

The third and last flaw we investigate deals with the scaling factor. While \cref{sec:dropout} and \cref{sec:landscape} considered it to be $1$ without loss of generality, we now use back the value from  \cref{def:lora}, i.e., using $\eta = \frac{\alpha}{r}$ as the \textit{Scaling Factor}.

The scaling factor plays an important role to match $||\mB\mA||$ to a comparable level with $||\mW||$. Despite its effectiveness, the scaling factor creates a ripple effect across layers of a LLM and may make finetuning unstable. \cite{hu2021lora} discussed the importance of the scaling factor and suggest to tune it carefully to prevent $\mB\mA$ from overwhelming $\mW$. From a different perspective, \cite{houlsby2019parameter} empirically showed the scale of the initialization of $\mB\mA$ can negatively impact validation accuracy. Later, \cite{hayou2024impact} argued that for the best performance, either $\mA$ or $\mB$ must be initialized at $0$. 

To illustrate the ripple effect, we adopt a multi-linear model which is a simplified version of the toy model in \cite{hayou2024lora+}.
\begin{equation}
   f_l(\vx) = \mW_l f_{l-1}(\vx),\quad l\in [L] 
   \label{eq:layered}
\end{equation}
where $L \ge 1$ is the number of layers. Applying LoRA at each layer gives
$$ f_l(\vx) = (\mW_l + \eta \mB_l \mA_l)f_{l-1}(\vx)$$
Expanding the equation, we have
$ f_L(\vx) = (\mW_L + \eta \mB_L \mA_L)...(\mW_1 + \eta \mB_1 \mA_1)\vx$.
Let $||\cdot||_M$ be a matrix norm induced by a vector norm $||\cdot||_v$. Applying Cauchy-Schwartz and the triangle inequality, we have
% $$ ||f_L(\vx)||_v = ||(\mW_L + \eta \mB_L \mA_L)...(\mW_1 + \eta \mB_1 \mA_1)||_M \cdot ||\vx||_v$$
% We abuse the notation to use $||\cdot||$ for both matrix norm and vector norm, applying triangle inequality, we have
\begin{align*}
  ||f_L(\vx)|| &\le ||f_L(\vx)||_v = ||(\mW_L + \eta \mB_L \mA_L)...(\mW_1 + \eta \mB_1 \mA_1)||_M \cdot ||\vx||_v\\
  & \le (||\mW_L|| + \eta ||\mB_L \mA_L||) ... (||\mW_1|| + \eta ||\mB_1 \mA_1||) ||\vx|| \\
 & = C (1 + \eta \frac{||\mB_L \mA_L||} {||\mW_L||})...(1 + \eta \frac{||\mB_1 \mA_1||}{||\mW_1||})||\vx|| \\
 & \le C (1 + \eta \bar{m})^L ||\vx|| = \Theta((1+\eta)^L)
\end{align*}
where $C = ||\mW_L||...||\mW_1||$ is a constant, and $\bar{m} = \frac{1}{L}\sum_{l\in L}{\frac{||\mB_l \mA_l||}{||\mW_L||}}$ is also a constant in a single forward pass. Notice that all the inequalities are tight.

\begin{proposition} (Ripple Effect) In the worst case, a constant scaling factor $\eta$ may cause the final output of a single forward pass of a LoRA finetuned model to grow exponentially \textit{w.r.t.} the number of layers in the model.
    \label{prop:ripple}
\end{proposition}

We also note that \cref{prop:ripple} is especially limiting for LLMs that most commonly resort to increased depth rather than increased width to scale up their capacity \citep{hestness2017deep}.

\begin{figure}[t!]
    \centering
    \includegraphics[width=\linewidth]{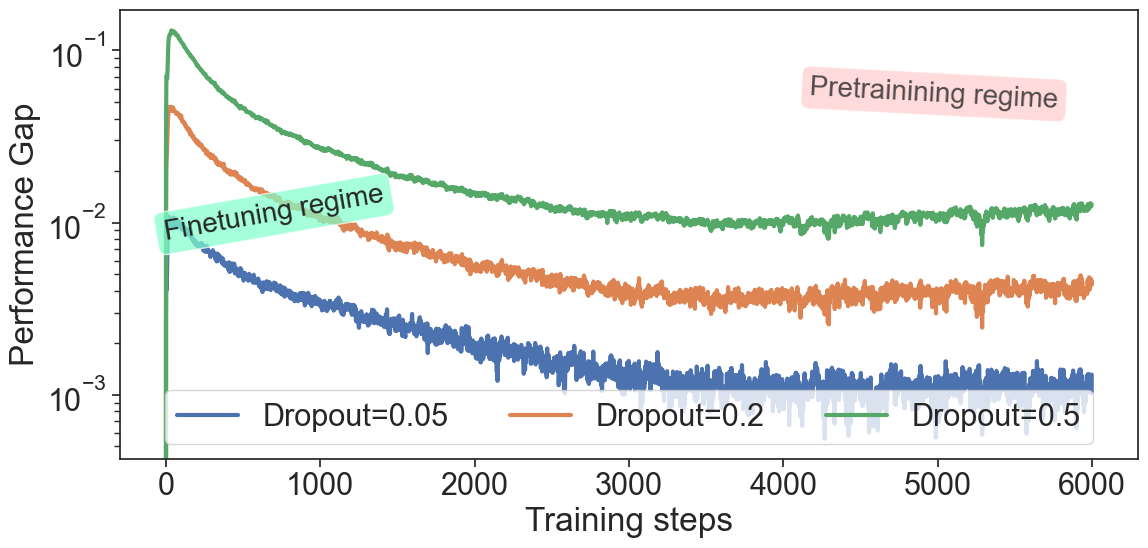}
    \caption{Test set performance gap ({\bf y-axis}) between the close-form Dropout regularization and its empirical estimate as a function of training steps ({\bf x-axis}). We observe that {\bf the benefit of Dropout as a regularizer falls short for finetuning (small number of training steps) compared to pretraining regimes (large number of trainign steps)}.}
    \label{fig:lora_gap}
\end{figure}
% \thisfloatsetup{subfloatrowsep=none}
\begin{figure}[t!]
% \begin{floatrow}
\centering
  \includegraphics[width=0.49\linewidth]{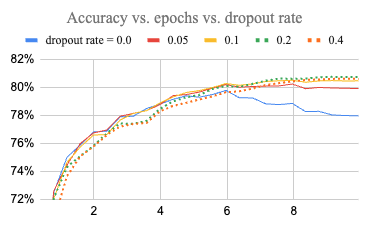}
  \includegraphics[width=0.49\textwidth]{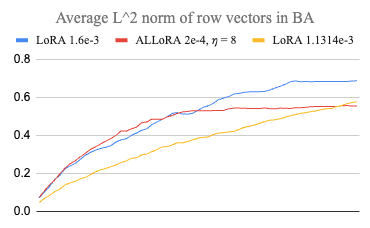}
\caption{\small \textbf{Left}: LoRA with varying Dropout rates: {\bf High value of Dropout provides the strongest performance after long fine-tuning and the weakest performance after short fine-tuning}. Each line is an average of 3 runs. X-axis is epochs, and Y-axis is accuracy. \textbf{Right}: ALLoRA escapes from 0 rapidly, and then tapers off into a measured move. The starting phase matches that of LoRA with a much higher learning rate. LoRA with a lower learning rate can reach the same level of $L^2$ norm but much slower. This finding echoes \cref{fig:convergence_lora} that showed how Dropout's induced noise does not converge until long training is employed.}
\label{fig:accuracy-epoch-dropout}
\end{figure}

\subsection{Empirical validation of the harmful impact of Dropout for short fine-tuning}
\label{sec:validation}

Because we were able to derive the expectation in close-form, we can now perform LoRA fine-tuning on the expected loss to see how the idealised performance varies. To that end, we propose a simple experiment with a 3-layer MLP on MNIST. We pretrain the model on a subset of $4096$ training set images, and then perform LoRA fine-tunining on another set of $512$ training images. We measure the gap in the test loss when training on the idealized (expected) loss and on the empirical loss on the test set throughout finetuning in \cref{fig:lora_gap}. We clearly observe that not only the gap consistently increases with the Dropout rate, but also that the gap culminates after a few hundred training steps and then slowly goes down. In short, the detrimental impact of Dropout is maximum during short finetuning episodes.

Moving to LLMs, we also confirm that simplified model's intuition with LLM experiments below. We run experiments with \textbf{Qwen2-0.5B} on \textbf{Bias in Bios}, a classification task to predict the job of an employee given the job description, for various dropout rates
and up to 10 epochs. We note that 10 epochs is already a large number of finetuning iterations for practical scenarios. It shows that large dropout rates successfully avoid overfitting, but at the cost of lower accuracy at lower number
of epochs, while no dropout sees higher accuracy at lower number of epochs, but will overfit at later epochs. See \cref{fig:accuracy-epoch-dropout} Left and \cref{tab:accuracy-epoch-dropout}.

Having concluded our brief tour of LoRA's possible shortcomings when it comes to fine-tuning LLMs in a few shots, we now propose to study our attempt at improving LoRA through a novel parametrization.

\section{ALLoRA: Escaping LoRA's Flaws for Fine-Tuning}
\label{sec:allora}

\subsection{Deconstructing LoRA}
\Cref{sec:flaw} summarized three flaws of LoRA, which we show can be addressed by a single solution: ALLoRA.

First we establish the underlying links among dropout, scaling factor, and learning rate. Consider the LoRA finetuning of a single layer as in \cref{eq:layered}, $f(\vx) = (\mW + \eta \mB \mA)\vx$. Following \cite{hayou2024lora+} and without loss of generality, we can simplify the model by assuming $\mW=0$, which is equivalent to defining $\tilde{\vy} = \vy - \mW \vx$, and rewriting the loss function by $\tilde{\vy}$. Also assuming $\eta = 1$, we have $f(\vx) = \mB\mA\vx$. The goal is to minimize loss $\mathcal{L}$ whose gradient is $g = \frac{\partial \mathcal{L}}{\partial (\mB\mA)}$. $f(\vx) \in \R^{n_1}$ is a column vector. Expanding it per row gives
\begin{equation}
    (f(\vx))_i = (\mB\mA)_{i, :} \vx,\quad i \in [n_1]
    \label{eq:single}
\end{equation}

The effect of dropping out $(f(\vx))_i$ for a given $i$ is equivalent to applying a per-row scaling factor $\eta_i = 0$ to $(\mB\mA)_{i, :}$. Note that this is true only for $(\mB\mA)_{i, :}$, the effect on $(\mB\mA)_{j, :}, j\ne i$ is slightly different. Since $\frac{d\eta \cdot f(x)}{dx} = \eta\frac{df(x)}{x}$, $\eta_i=0$ is implicitly applied to the $i$-th row of the gradient $g_i = (\frac{\partial \mathcal{L}}{\partial (\mB\mA)})_{i, :}$, which is again a scaling factor applied to the learning rate $l$.

The observation reveals that both scaling factor and dropout are adaptions on LoRA output $f(x)$, and both have effects on gradient. We are inspired to formalize a general framework that subsumes both, within which we can use a principled approach to systematically discover novel solutions.
% Inspired by this observation, our intuition is to adapt learning rate to $(BA)_{i, :}$ such that small ones grow rapidly while large ones slow down to discover optimal path.

\begin{definition} (Adaptive Learning) Consider a single layer linear model $f(\vx) = \mB\mA\vx$ with gradient
$g(\vx) = \frac{\partial \mathcal{L}(f(\vx))}{\partial (\mB\mA)}$. Let \textit{Output Adaptor} be a function $f_o: \R^{n_1} \rightarrow \R^{n_1}$, \textit{Gradient Adaptor} be a function
$f_g: \R^{n_1 \times n_2} \rightarrow \R^{n_1 \times n_2}$. Define adapted output $\tilde{f}$ and adapted gradient $\tilde{g}$ by
\begin{equation}
\begin{cases}
    \tilde{f} = f_o \circ f \\
    \tilde{g} = f_g \circ g
\end{cases}
\end{equation}
Adaptive learning is to use the adapted $\tilde{f}$ and $\tilde{g}$ in place of $f$ and $g$ respectively in the learning process.
    \label{def:adapt}
\end{definition}

Let $I: x \mapsto x$ be the identity function. Then a natural corollary is that all learning is adaptive learning (when $f_o = I$ and $f_g = I$). Note that $\mathcal{L}$ is a function of $f(\vx)$, hence $g(\vx) = \frac{\partial (\mathcal{L} \circ f)}{\partial (\mB\mA)}$. Use $\tilde{f}$ in place of $f$ defines a naturally adapted gradient $\tilde{g} = \frac{\partial (\mathcal{L} \circ \tilde{f})}{\partial (\mB\mA)} = \frac{\partial (\mathcal{L} \circ f_o \circ f)}{\partial (\mB\mA)}$. When it's clear from the context, we omit $\tilde{g}$ if it's naturally defined by a non-trivial $\tilde{f}$.
% It's straightforward to define $f_o$ and $f_g$ for scaling factor and dropout, see \ref{sec:ale} for details.

\subsection{ALLoRA: Less Hyper-Parameters and More Stability for Finetuning}
% \label{sec:allora}

Under the Adaptive Learning framework, scaling factor is define by $f_o = \kappa: x \mapsto \eta x$.
\begin{equation}
    \begin{cases}
    \tilde{f} = \kappa \circ f = \eta f \\
    \tilde{g} = \kappa \circ f = \eta g
\end{cases}
\end{equation}

One idea to reduce the ripple effect while keeping the positive effect of scaling factor is to force $f_o = I$, while keeping $\tilde{g}$ intact, which is to use a larger learning rate $\eta \cdot l$. Nonetheless, a fixed learning rate cannot simultaneously achieve both fast escape from 0 and, once away from 0, measured discovery of optimal direction. We think an adaptive learning rate that is inversely proportional to $||(\mB\mA)_{i, :}||$ is a good candidate to realize our idea. We use the function $1 / \sqrt{||(\mB\mA)_{i, :}|| + 1 / \eta^2}$ which reaches maximum $\eta$ at $||(\mB\mA)_{i, :}||=0$ and then tapers down when $||(\mB\mA)_{i, :}||$ increases (\cref{fig:adapt-lr}).
% \subsection{Adaptive Learning Rate}\label{sec:alr}
% Following Definition \ref{def:adapt}, scaling factor can be defined by $f_o: x \mapsto \eta x$, which applies a constant $\eta$ universally. Intuitively, for the purpose of scaling up $||BAx||$ to be comparable with $||W^\ast x||$, it is more appropriate to apply a large scaling factor when $(BA)_{i, :}$ is small and a smaller one when $(BA)_{i, :}$ is large, such that small ones grow rapidly while large ones slow down to discover optimal directions.

% Proposition \ref{prop:sflr} implies that for each $f_o$, there is a corresponding $f_g$ that applies the same effect on gradient. One idea is to only keep $f_g$ and force $f_o = I$, in which way there will be no ripple effect in the forward pass as output from lower layer to higher layer is unchanged. Thus, instead of an adaptive scaling factor on output, a better solution might be an adaptive scaling factor on gradient, hence the name of \textit{Adaptive Learning Rate}.

% We think a function that is inversely proportional to $||(BA)_{i, :}||$ is a good candidate to realize our idea. We use the function $1 / \sqrt{||(BA)_{i, :}|| + 1 / \eta^2}$ which reaches maximum at $||(BA)_{i, :}||=0$ and then tapers down when $||(BA)_{i, :}||$ increases (Figure \ref{fig:scale-factor-learing-rate} Right).

Formally, ALLoRA is defined by
$$\begin{cases}
    \tilde{f} = I \circ f \\
    \tilde{g}_i = 1 / \sqrt{||(\mB\mA)_{i, :}|| + 1 / \eta^2} \cdot g_i,\quad i \in [n_1]
\end{cases}$$
where $\eta$ is a hyperparameter. Note that this does not introduce a new hyperparameter. We split
learning rate into a constant base learning rate $l_b$ and $\eta$, and the effective learning rate is $\eta \cdot l_b$.

One more implementation detail is the backward pass computes, in addition to the gradients of $\mA$ and $\mB$, also the gradient of the input from layer below, and propagate which back to the layer below. We only modify the gradients of $\mA$ and $\mB$, but not that of the input. This helps further restrict the changes within each layer and reduce ripple effect.

% \begin{figure}
%     \centering
%     \begin{tabular}{cc}
%     \includegraphics[width=0.45\textwidth]{fig-average-L2-norm.png} & \includegraphics[width=0.45\textwidth]{fig-average-angle-change.png} \\
%     \end{tabular}
%     \caption{\textbf{Left}: The $L^2$-norm of row vectors in $BA$ grows faster under ALLoRA than under LoRA.
%     \textbf{Right}: After a few steps at the beginning, ALLoRA sees greater angle change of row vectors in $BA$ than LoRA does.}
%     \label{fig:row-vector}
% \end{figure}

To quickly verify our idea, we add probing code to trace the $L^2$ norm of row vectors of $\mB\mA$. As shown in \cref{fig:accuracy-epoch-dropout} Right, adaptive learning rate escapes from $0$ rapidly, the speed matches with LoRA with a learning rate that is $\eta \cdot l_b$. Then it finds an approriate level and enters measured discovery of optimal directions. LoRA with a learning rate lower than $\eta \cdot l_b$ can reach the same level, but at a much slower pace. The experiment is with \textbf{Snowflake Arctic XS} and \textbf{Rotten Tomatoes}.

Note that ALLoRA multiplies different scaling factors to different rows of $\mB\mA$'s gradient stochastically (because $\mA$ is stochastically initialized and $\mB\mA$ is stochastically learnt). Intuitively it is a generalization of Dropout which multiplies binary factors ($0$ or $\frac{1}{1-p}$) to different rows stochastically. We hypothesize that ALLoRA may recover some regularisation effect of Dropout and invite researchers to find a theoretical proof.

% To quickly verify our idea, we add probing code to trace the $L^2$-norm and angle changes of row vectors of $BA$. To be more specific, let $v^k_i$ be the $i$-th row vector of $BA$ at training step $k$, we track $||v^k_i||$ and $||v^k_i - \frac{v^k_i \cdot v^{k-1}_i}{||v^{k-1}_i||}||$. As shown in Figure \ref{fig:row-vector}, adaptive learning rate sees higher $||v^k_i||$ and, after a few initial steps, higher angle changes, indicating that ALLoRA grows $BA$ more rapidly and spend more training to discover optimal directions. The experiment is with \textbf{Snowflake Arctic XS} and \textbf{Rotten Tomatoes}.

% Worth noting is that in the experiment, ALLoRA uses $l_b=1e-4, \eta=2$. To be fair, LoRA uses $l=2e-4$, hence they have the same learning rate when training step $k=0$. Even though at $k>0$, ALLoRA's gradient is scaled by a factor strictly less than $2$, it still leads LoRA in both the $L^2$-norm and angle changes. The interesting finding motivated us to try trivial $\eta=1$ which also admits good results in our experiments.

\subsection{A Family of Adaptive Solutions}
We adopt a principled approach to explore other reasonable designs that fall into the adaptive learning framework defined by  \cref{def:adapt}.

First notice that in $(f(\vx))_i = (\mB\mA)_{i, :} \vx$, $(f(\vx))_i$ and $(\mB\mA)_{i, :}$ define each other. So instead of adapt the learning by $(\mB\mA)_{i, :}$,
we can also adapt it by $(f(\vx))_i$, which is \textit{Output-Dependent}, or ALLoRA-OD, defined by
$$\begin{cases}
    \tilde{f} = I \circ f \\
    \tilde{g}_i = 1 / \sqrt{|(f(\vx))_i| + 1 / \eta^2} \cdot g_i,\quad i\in [n_1]
\end{cases}$$
Note that $(f(\vx))_i$ is a scalar, hence we use its absolute value. Qualitatively, ALLoRA-OD subjects to the stochastic noise in $\vx$ because $(f(\vx))_i = (\mB\mA)_{i, :} \vx$. According to \cite{smith2021origin}, this type of stochastic noise is an implicit regularization. Our conjecture is that it may drag down the accuracy just as dropout does, and therefore ALLoRA-OD may not be as good as ALLoRA.

Given the link between learning rate and scaling factor, we may achieve similar effect by switching from adaptive learning rate to \textit{Adaptive Scaling Factor}, or ASF-LoRA, defined by $$\tilde{f}_i = 1 / \sqrt{|(f(\vx))_i| + 1 / \eta^2} \cdot f_i,\quad i\in [n_1]$$
Note that $\tilde{g}$ is naturally defined by using $\tilde{f}$ in place of $f$.
The potential downside is that it introduces ripple effect across layers, which may blur accuracy. And our conjecture is again ASF-LoRA may not be as good as ALLoRA.

One more caveat of ASF-LoRA is that we cannot merge $\mB\mA$ back to $\mW$ as we need to apply $f_o$ to the LoRA output.

\subsection{Empirical Validation: Perception Tasks}\label{sec:percept}
Our first set of experiments gauges the performance of ALLoRA on perception tasks. Mainstream LLMs nowadays are mostly pretrained by next token prediction, which is good for generative tasks, but may not be a good fit for perception tasks such as Natural Language Understanding (NLU) and Sentiment Analysis (SA). In fact, we observe subpar accuracy when finetuning popular open-weight models for NLU and SA tasks (see \cref{tab:nlu_full}). We hope to show that ALLoRA may help boost the accuracy for perception tasks.

For our experiments, we pick three midsized LLMs: \textbf{Qwen2-0.5B}, \textbf{Snowflake-Artic-L}, and \textbf{OpenELM-450M}, and four NLU and SA datasets: \textbf{Bias in Bios}, \textbf{Emotion}, \textbf{Rotten Tomatoes}, and \textbf{Yelp Review}. To demonstrate the stability of ALLoRA, we run the experiments with various $\eta^2 \in \{1, 2, 4\}$ with a fixed baseline learning rate $l_b=1e-4$. To be fair for LoRA, we run LoRA at learning rate $l \in \{1e-4, \sqrt{2}e-4, 2e-4\}$, respectively. We finetune for $2$ epochs. Each experiment is run $5$ times and we report average final accuracy. \Cref{fig:nlu} Left shows the accuracy gap between ALLoRA and LoRA, where positive numbers indicate ALLoRA has better accuracy. The result shows that ALLoRA in general admits better accuracy over plain LoRA. Average improvement over all cases is $0.3\%$.

In the experiment, the dropout rate for ALLoRA is $0.0$ and that for LoRA is $0.05$. We also run ALLoRA+D, the version of ALLoRA with \textbf{D}ropout, also at dropout rate $0.05$. \Cref{fig:nlu} Right shows that there is no evident difference between ALLoRA and ALLoRA+D, matching our theoretical result from \cref{sec:dropout}.

Since $\eta$ originates from $\frac{\alpha}{r}$ in \cref{def:lora}, we also run at various LoRA ranks $r$, and the results show that ALLoRA's advantage is consistent across different $r$ (\cref{tab:rank}).

\begin{table}[t!]
\setlength{\tabcolsep}{3pt}
    \centering
    \caption{\small Accuracy comparison of various LoRA ranks on \textbf{Qwen2-0.5B} and \textbf{Emotion}.
     LoRA's and DoRA's learning rate $l=1e-4$, ALLoRA's base learning rate is $l_b=1e-4$, and $\eta=1$. Each cell is an average over 5 runs.}
    \label{tab:rank}
    \small
\begin{tabular}{cccccc}
  \toprule
  Method & $r=4$ & $r=8$ & $r=16$ & $r=32$ & $r=64$ \\
  \midrule
  LoRA & 33.09 & 34.01 & 35.19 & 35.59 & 38.13 \\
  DoRA & 33.34 & 34.14 & 35.50 & 36.81 & 38.07 \\
  ALLoRA & \textbf{33.45} & \textbf{34.80} & \textbf{35.61} & \textbf{37.18} & \textbf{38.27} \\
  \bottomrule
\end{tabular}
\end{table}

\begin{table}[t!]
    \centering
    \caption{\small Accuracy gap between ALLoRA and LoRA. Each cell is an average of 5 runs.
    \textbf{Left}: ALLoRA admits better accuracy than that of LoRA.
    \textbf{Right}: ALLoRA+D, the version with $0.05$ dropout rate, admits comparable accuracy than that of ALLoRA.}
    \label{fig:nlu}
    \begin{tabular}{cc}
    \includegraphics[width=0.45\textwidth]{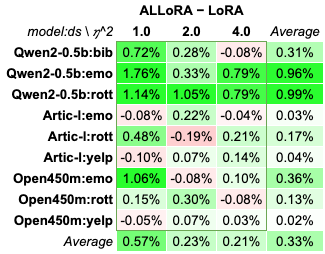} & \includegraphics[width=0.45\textwidth]{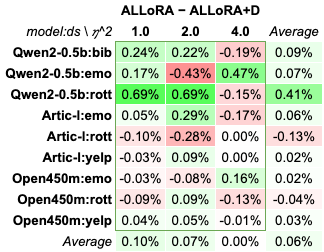} \\
    \end{tabular}
\end{table}

\subsection{Empirical Validation: Commonsense Reasoning}
\label{sec:commonsense}

We also compare ALLoRA with DoRA (\cite{liu2024dora}), a recent LoRA variant that demonstrated superb performance over a range of PEFT methods. Since DoRA results are universally better than other PEFT methods, we only compare ALLoRA to DoRA. We run experiments on \textbf{LLaMA-7B}, \textbf{LLaMA2-7B}, and \textbf{LLaMA3-8B} on 8 \textbf{Commonsense} tasks. Following DoRA’s setup, for each model, we run both ALLoRA and ALLoRA+D with LoRA rank $r \in \{16, 32\}$ and for $3$ epochs. \Cref{tab:cs} shows that for all cases, either ALLoRA or ALLoRA+D has the best average accuracy. On average, ALLoRA and ALLoRA+D each boosted accuracy by $0.3\%$ over DoRA.

Note that we run experiments with various $\eta^2 \in \{1, 2, 4\}$ and report the best accuracy, this follows DoRA’s practices to run with various learning rate $l \in \{1e-4, 2e-4\}$ and report the best. 

In \cref{tab:cs} we also report the number of trainable parameters as a percentage of the number of pretrained parameters. Since ALLoRA does not introduce additional trainable parameters, its trainable parameters are slightly lower than that of DoRA.

\begin{table}
\setlength{\tabcolsep}{1pt}
    \centering
    \caption{\small Accuracy comparison of LLaMA 7B, LLaMA2 7B, and LLaMA3 8B between ALLoRA and DoRA on eight commonsense
reasoning datasets. DoRA results are taken from \cite{liu2024dora}. ALLoRA+D is the version of ALLoRA with $0.05$ dropout rate.}
    \label{tab:cs}
    \scriptsize
\begin{tabular}{cccccccccccc}
\toprule
\begin{tabular}{c}Model \\ LoRA Rank \end{tabular} 
               & Method & \begin{tabular}{c}\# Params \\ \% \end{tabular}
               & BoolQ & PIQA & SIQA & HSwag & WGrande & ARC-e & ARC-c & OBQA & Avg. \\
\midrule
\multirow[m]{3}{*}{\begin{tabular}{c}LLaMA-7B \\ 16\end{tabular}} 
               & DoRA &  0.43 & 70.0 & 82.6 & 79.7 & 83.2 & 80.6 & 80.6 & 65.4 & 77.6 & 77.5 \\
               & ALLoRA+D (ours) & 0.41 & 69.4 & 82.7 & 78.3 & 84.8 & 80.0 & 80.9 & 65.7 & 79.2 & \textbf{77.6} \\
               & ALLoRA (ours) & 0.41 & 69.2 & 80.8 & 78.5 & 83.9 & 81.1 & 80.8 & 65.2 & 78.2 & 77.2 \\
\midrule
\multirow[m]{3}{*}{\begin{tabular}{c}LLaMA-7B \\ 32\end{tabular}}
               & DoRA & 0.84 & 69.7 & 83.4 & 78.6 & 87.2 & 81.0 & 81.9 & 66.2 & 79.2 & 78.4 \\
               & ALLoRA+D (ours) & 0.83 & 70.0 & 82.3 & 78.1 & 84.6 & 82.2 & 81.0 & 67.9 & 81.0 & \textbf{78.4} \\
               & ALLoRA (ours) & 0.83 & 70.2 & 82.6 & 78.6 & 83.8 & 81.1 & 81.0 & 66.3 & 82.6 & 78.3 \\
\midrule
\multirow[m]{3}{*}{\begin{tabular}{c}LLaMA2-7B \\ 16\end{tabular}}
               & DoRA & 0.43 & 72.0 & 83.1 & 79.9 & 89.1 & 83.0 & 84.5 & 71.0 & 81.2 & 80.5 \\
               & ALLoRA+D (ours) & 0.41 & 71.7 & 83.7 & 79.5 & 91.4 & 82.4 & 84.3 & 69.2 & 81.2 & 80.4 \\
               & ALLoRA (ours) & 0.41 & 72.4 & 83.9 & 80.0 & 90.8 & 83.0 & 84.7 & 71.3 & 80.2 & \textbf{80.8} \\
\midrule
\multirow[m]{3}{*}{\begin{tabular}{c}LLaMA2-7B \\ 32\end{tabular}}
               & DoRA &  0.84 & 71.8 & 83.7 & 76.0 & 89.1 & 82.6 & 83.7 & 68.2 & 82.4 & 79.7 \\
               & ALLoRA+D (ours) & 0.83 & 72.2 & 83.1 & 79.6 & 91.2 & 84.5 & 84.5 & 71.0 & 80.0 & 80.8 \\
               & ALLoRA (ours) & 0.83 & 72.3 & 83.8 & 79.3 & 91.4 & 83.0 & 85.0 & 71.2 & 82.2 & \textbf{81.0} \\
\midrule
\multirow[m]{3}{*}{\begin{tabular}{c}LLaMA3-8B \\ 16\end{tabular}}
               & DoRA & 0.35 & 74.5 & 88.8 & 80.3 & 95.5 & 84.7 & 90.1 & 79.1 & 87.2 & 85.0 \\
               & ALLoRA+D (ours) & 0.35 & 75.2 & 88.9 & 80.8 & 95.6 & 84.7 & 90.2 & 80.6 & 85.8 & 85.2 \\
               & ALLoRA (ours) & 0.35 & 74.5 & 89.1 & 80.4 & 95.5 & 85.8 & 90.7 & 80.3 & 86.0 & \textbf{85.3} \\
\midrule
\multirow[m]{3}{*}{\begin{tabular}{c}LLaMA3-8B \\ 32\end{tabular}}
               & DoRA & 0.71 & 74.6 & 89.3 & 79.9 & 95.5 & 85.6 & 90.5 & 80.4 & 85.8 & 85.2 \\
               & ALLoRA+D (ours) & 0.70 & 74.5 & 88.9 & 81.8 & 95.9 & 86.3 & 90.4 & 80.5 & 87.6 & \textbf{85.8} \\
               & ALLoRA (ours) & 0.70 & 75.1 & 88.7 & 81.8 & 95.8 & 85.4 & 91.0 & 81.1 & 86.6 & 85.7 \\
\bottomrule
\end{tabular}
\end{table}

\begin{table}[t!]
    \centering
    \caption{\small Accuracy gap between ALLoRA and other adaptive approaches. Each cell is an average of 5 runs.
    \textbf{Top left}: ALLoRA admits better accuracy than that of ALLoRA-OD, where learning rate is LoRA \textbf{O}utput-\textbf{D}ependent.
    \textbf{Top right}: ALLoRA admits better accuracy than that of ASF-LoRA, where an \textbf{A}daptive \textbf{S}cale \textbf{F}actor
    is applied to LoRA output.
    \textbf{Bottom left}: ALLoRA admits better accuracy than that of LoRA with a fixed scaling factor.
    \textbf{Bottom right}: Adaptive scale factor is not better than a fixed scale factor.}
    \label{fig:ablation}
    \begin{tabular}{cc}
    \includegraphics[width=0.45\textwidth]{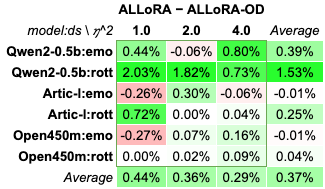} & \includegraphics[width=0.45\textwidth]{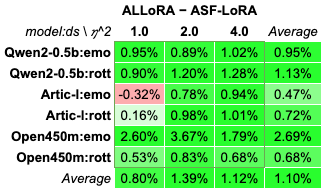} \\
    \includegraphics[width=0.45\textwidth]{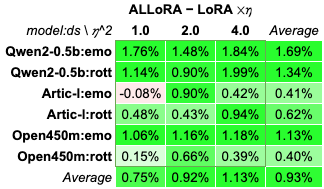} & \includegraphics[width=0.45\textwidth]{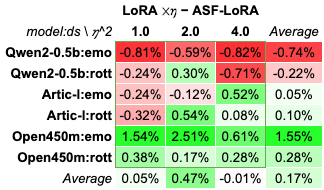} \\
    \end{tabular}
\end{table}
\section{Ablation Study}
Using the same setup in \cref{sec:percept}, we run experiments with ALLoRA-OD, the output-dependent variant, and ASF-LoRA, the adaptive scaling factor variant. We also run LoRA with comparable fixed scaling factors to form an objective baseline for ASF-LoRA.

\subsection{ALLoRA-OD}
\Cref{fig:ablation} Top Left shows the accuracy gap between ALLoRA and ALLoRA-OD. A positive number indicates that ALLoRA has better accuracy. Overall speaking, ALLoRA has better accuracy than ALLoRA-OD. But the difference is moderate, as average improvement over all cases is $0.4\%$.

The result matches our conjecture that stochastic noise experienced by ALLoRA-OD might have dragged down accuracy at early epochs. 

\subsection{ASF-LoRA and LoRA with Fixed Scaling Factor}
Since a scaling factor on output is implicitly also a scaling factor on gradient, we use the same $\eta$ when comparing between ALLoRA and ASF-LoRA, i.e., the gradient adaptor $f_g$ in ALLoRA and the output adaptor $f_o$ in ASF-LoRA use the same $\eta$.

\Cref{fig:ablation} Top Right shows the accuracy gap between ALLoRA and ASF-LoRA. A positive number indicates that ALLoRA has better accuracy. ALLoRA has a significant advantage over ASF-LoRA as average improvement over all cases is $1.1\%$. Since we know that ALLoRA-OD is only slightly worse than ALLoRA, the evidence leans toward that Adaptive Learning Rate is in general a better solution family than Adaptive Scaling Factor.

We also run LoRA at comparable fixed scaling factors $\frac{\alpha}{r} \in \{1, \sqrt{2}, 2 \}$. The results, as shown in \cref{fig:ablation} Bottom, show that
\begin{itemize}
    \item ASF-LoRA is not a competitive method, as over half of cases see ASF-LoRA's accuracy significantly lower than LoRA with a comparable fixed scaling factor (positive numbers in \cref{fig:ablation} Bottom Right).
    \item ALLoRA is significantly better than LoRA with a comparable fixed scaling factor, as average improvement over all cases is $0.9\%$.
\end{itemize}

\section{Conclusion and Future Work}
This paper identifies three major flaws of LoRA, namely dropout, zero-initialization, and scaling factor. We conducted principled analysis and proved that dropout is not a must-have in the finetuning regime. After uncovering the hidden connection between dropout, scaling factor, and learning rate, we proposed a unified adaptive learning framework to address them all: ALLoRA. Empirical results show that ALLoRA admits better accuracy than plain LoRA over multiple backbones, datasets, and learning rates; and better accuracy than recent successful LoRA variants such as DoRA. Ablation study shows that ALLoRA is the optimal in a family of adaptive methods.

We list a few interesting research directions and invite researchers to explore the frontier opened-up by our research:
\begin{itemize}
    \item The adaptive learning framework introduced by this paper is generic and may find broad applications beyond LoRA. Other use cases such as pretraining may not have the constraints that the weight matrix must be initialized at $0$. But they may have other types of constraints, which may be solved by adaptive learning with a different adaptor function.
    \item Within the LoRA use case, we only examine one particular adaptor function, there could be other adaptor functions that have superior performance.
    \item We only provide empirical evidence that ALLoRA admits better accuracy and hypothesize that ALLoRA is implicitly a regularization. Theoretical guarantee is needed, especially for the convoluted case where the base model has multiple layers.
    \item Starting from $0$ weights may avoid the lottery ticket hypothesis (\cite{frankle2019lottery}), for good or bad, where adaptive learning rate can be a handy tool.
\end{itemize}

\bibliography{allora_iclr2025.bib}

\begin{thebibliography}{37}
\providecommand{\natexlab}[1]{#1}
\providecommand{\url}[1]{\texttt{#1}}
\expandafter\ifx\csname urlstyle\endcsname\relax
  \providecommand{\doi}[1]{doi: #1}\else
  \providecommand{\doi}{doi: \begingroup \urlstyle{rm}\Url}\fi

\bibitem[Chowdhary(2020)]{chowdhary2020natural}
KR~Chowdhary.
\newblock Natural language processing.
\newblock \emph{Fundamentals of artificial intelligence}, pp.\  603--649, 2020.

\bibitem[Dubey et~al.(2024)Dubey, Jauhri, Pandey, Kadian, Al-Dahle, Letman, Mathur, Schelten, Yang, Fan, et~al.]{dubey2024llamaf}
Abhimanyu Dubey, Abhinav Jauhri, Abhinav Pandey, Abhishek Kadian, Ahmad Al-Dahle, Aiesha Letman, Akhil Mathur, Alan Schelten, Amy Yang, Angela Fan, et~al.
\newblock The llama 3 herd of models.
\newblock \emph{arXiv preprint arXiv:2407.21783}, 2024.

\bibitem[Frankle \& Carbin(2019)Frankle and Carbin]{frankle2019lottery}
Jonathan Frankle and Michael Carbin.
\newblock The lottery ticket hypothesis: Finding sparse, trainable neural networks.
\newblock In \emph{International Conference on Learning Representations}, 2019.

\bibitem[Hayou et~al.(2024{\natexlab{a}})Hayou, Ghosh, and Yu]{hayou2024impact}
Soufiane Hayou, Nikhil Ghosh, and Bin Yu.
\newblock The impact of initialization on lora finetuning dynamics.
\newblock \emph{arXiv preprint arXiv:2406.08447}, 2024{\natexlab{a}}.

\bibitem[Hayou et~al.(2024{\natexlab{b}})Hayou, Ghosh, and Yu]{hayou2024lora+}
Soufiane Hayou, Nikhil Ghosh, and Bin Yu.
\newblock Lora+: Efficient low rank adaptation of large models.
\newblock \emph{arXiv preprint arXiv:2402.12354}, 2024{\natexlab{b}}.

\bibitem[He et~al.(2022)He, Zhou, Ma, Berg-Kirkpatrick, and Neubig]{he2022towards}
Junxian He, Chunting Zhou, Xuezhe Ma, Taylor Berg-Kirkpatrick, and Graham Neubig.
\newblock Towards a unified view of parameter-efficient transfer learning.
\newblock In \emph{International Conference on Learning Representations}, 2022.

\bibitem[Hestness et~al.(2017)Hestness, Narang, Ardalani, Diamos, Jun, Kianinejad, Patwary, Yang, and Zhou]{hestness2017deep}
Joel Hestness, Sharan Narang, Newsha Ardalani, Gregory Diamos, Heewoo Jun, Hassan Kianinejad, Md~Mostofa~Ali Patwary, Yang Yang, and Yanqi Zhou.
\newblock Deep learning scaling is predictable, empirically.
\newblock \emph{arXiv preprint arXiv:1712.00409}, 2017.

\bibitem[Hoffmann et~al.(2022)Hoffmann, Borgeaud, Mensch, Buchatskaya, Cai, Rutherford, Casas, Hendricks, Welbl, Clark, et~al.]{hoffmann2022training}
Jordan Hoffmann, Sebastian Borgeaud, Arthur Mensch, Elena Buchatskaya, Trevor Cai, Eliza Rutherford, Diego de~Las Casas, Lisa~Anne Hendricks, Johannes Welbl, Aidan Clark, et~al.
\newblock Training compute-optimal large language models.
\newblock \emph{arXiv preprint arXiv:2203.15556}, 2022.

\bibitem[Houlsby et~al.(2019)Houlsby, Giurgiu, Jastrzebski, Morrone, De~Laroussilhe, Gesmundo, Attariyan, and Gelly]{houlsby2019parameter}
Neil Houlsby, Andrei Giurgiu, Stanislaw Jastrzebski, Bruna Morrone, Quentin De~Laroussilhe, Andrea Gesmundo, Mona Attariyan, and Sylvain Gelly.
\newblock Parameter-efficient transfer learning for nlp.
\newblock In \emph{International conference on machine learning}, pp.\  2790--2799. PMLR, 2019.

\bibitem[Hu et~al.(2021)Hu, Shen, Wallis, AllenZhu, Li, Wang, Wang, and Chen]{hu2021lora}
Edward~J. Hu, Yelong Shen, Phillip Wallis, Zeyuan AllenZhu, Yuanzhi Li, Shean Wang, Lu~Wang, and Weizhu Chen.
\newblock Lora: Low-rank adaptation of large language models.
\newblock \emph{arXiv preprint arXiv:2106.09685}, 2021.

\bibitem[Hyeon-Woo et~al.(2022)Hyeon-Woo, Ye-Bin, and Oh]{hyeonwoo2022lowrank}
Nam Hyeon-Woo, Moon Ye-Bin, and Tae-Hyun Oh.
\newblock Fedpara: Low-rank hadamard product for communication-efficient federated learning.
\newblock In \emph{ICLR} \citet{hyeonwoo2022lowrank}.

\bibitem[Jang et~al.(2024)Jang, Lee, and Ryu]{jang2024lora}
Uijeong Jang, Jason~D Lee, and Ernest~K Ryu.
\newblock Lora training in the ntk regime has no spurious local minima.
\newblock \emph{arXiv preprint arXiv:2402.11867}, 2024.

\bibitem[Jiang et~al.(2023)Jiang, Sablayrolles, Mensch, Bamford, Chaplot, Casas, Bressand, Lengyel, Lample, Saulnier, et~al.]{jiang2023mistral}
Albert~Q Jiang, Alexandre Sablayrolles, Arthur Mensch, Chris Bamford, Devendra~Singh Chaplot, Diego de~las Casas, Florian Bressand, Gianna Lengyel, Guillaume Lample, Lucile Saulnier, et~al.
\newblock Mistral 7b.
\newblock \emph{arXiv preprint arXiv:2310.06825}, 2023.

\bibitem[Karimi~Mahabadi et~al.(2021{\natexlab{a}})Karimi~Mahabadi, Henderson, and Ruder]{karimi2021hyper}
Rabeeh Karimi~Mahabadi, James Henderson, and Sebastian Ruder.
\newblock Compacter: Efficient low-rank hypercomplex adapter layers.
\newblock In M.~Ranzato, A.~Beygelzimer, Y.~Dauphin, P.S. Liang, and J.~Wortman Vaughan (eds.), \emph{Advances in Neural Information Processing Systems}, volume~34, pp.\  1022--1035. Curran Associates, Inc., 2021{\natexlab{a}}.

\bibitem[Karimi~Mahabadi et~al.(2021{\natexlab{b}})Karimi~Mahabadi, Ruder, Dehghani, and Henderson]{karimi2021parameterefficient}
Rabeeh Karimi~Mahabadi, Sebastian Ruder, Mostafa Dehghani, and James Henderson.
\newblock Parameter-efficient multi-task fine-tuning for transformers via shared hypernetworks.
\newblock In \emph{Annual Meeting of the Association for Computational Linguistics}, 2021{\natexlab{b}}.

\bibitem[Kopiczko et~al.(2023)Kopiczko, Blankevoort, and Asano]{kopiczko2024vera}
Dawid~Jan Kopiczko, Tijmen Blankevoort, and Yuki~Markus Asano.
\newblock Vera: Vector-based random matrix adaptation.
\newblock \emph{CoRR}, abs/2310.11454, 2023.

\bibitem[Lester et~al.(2021)Lester, Al-Rfou, and Constant]{lester2021emnlp}
Brian Lester, Rami Al-Rfou, and Noah Constant.
\newblock The power of scale for parameter-efficient prompt tuning.
\newblock In  \citet{lester2021emnlp}, pp.\  3045--3059.

\bibitem[Liu et~al.(2024{\natexlab{a}})Liu, Wang, Yin, Molchanov, Wang, Cheng, and Chen]{liu2024dora}
Shih-Yang Liu, Chien-Yi Wang, Hongxu Yin, Pavlo Molchanov, Yu-Chiang~Frank Wang, Kwang-Ting Cheng, and Min-Hung Chen.
\newblock Dora: Weight-decomposed low-rank adaptation.
\newblock \emph{arXiv preprint arXiv:2402.09353}, 2024{\natexlab{a}}.

\bibitem[Liu et~al.(2024{\natexlab{b}})Liu, Qiu, Feng, Xiu, Xue, Yu, Feng, Liu, Heo, Peng, Wen, Black, Weller, and Schölkopf]{liu2024param}
Weiyang Liu, Zeju Qiu, Yao Feng, Yuliang Xiu, Yuxuan Xue, Longhui Yu, Haiwen Feng, Zhen Liu, Juyeon Heo, Songyou Peng, Yandong Wen, Michael~J. Black, Adrian Weller, and Bernhard Schölkopf.
\newblock Parameter-efficient orthogonal finetuning via butterfly factorization.
\newblock In \emph{ICLR} \citet{liu2024param}.

\bibitem[Mehta et~al.(2024)Mehta, Sekhavat, Cao, Horton, Jin, Sun, Mirzadeh, Najibi, Belenko, Zatloukal, and Rastegari]{mehtaOpenELMEfficientLanguage2024}
Sachin Mehta, Mohammad~Hossein Sekhavat, Qingqing Cao, Maxwell Horton, Yanzi Jin, Chenfan Sun, Iman Mirzadeh, Mahyar Najibi, Dmitry Belenko, Peter Zatloukal, and Mohammad Rastegari.
\newblock {OpenELM}: {An} {Efficient} {Language} {Model} {Family} with {Open} {Training} and {Inference} {Framework}.
\newblock \emph{arXiv.org}, April 2024.
\newblock URL \url{https://arxiv.org/abs/2404.14619v1}.

\bibitem[Ponti et~al.(2022)Ponti, Sordoni, Bengio, and Reddy]{ponti2022combining}
Edoardo~M Ponti, Alessandro Sordoni, Yoshua Bengio, and Siva Reddy.
\newblock Combining modular skills in multitask learning.
\newblock \emph{arXiv preprint arXiv:2202.13914}, 2022.

\bibitem[Qiu et~al.(2023)Qiu, Liu, Feng, Xue, Feng, Liu, Zhang, Weller, and Schölkopf]{qiu2023control}
Zeju Qiu, Weiyang Liu, Haiwen Feng, Yuxuan Xue, Yao Feng, Zhen Liu, Dan Zhang, Adrian Weller, and Bernhard Schölkopf.
\newblock Controlling text-to-image diffusion by orthogonal finetuning.
\newblock In  \citet{qiu2023control}.

\bibitem[Razdaibiedina et~al.(2023)Razdaibiedina, Mao, Khabsa, Lewis, Hou, Ba, and Almahairi]{razdaibiedina2023residual}
Anastasia Razdaibiedina, Yuning Mao, Madian Khabsa, Mike Lewis, Rui Hou, Jimmy Ba, and Amjad Almahairi.
\newblock Residual prompt tuning: improving prompt tuning with residual reparameterization.
\newblock In  \citet{razdaibiedina2023residual}, pp.\  6740--6757.
\newblock ISBN 978-1-959429-62-3.

\bibitem[Renduchintala et~al.(2023)Renduchintala, Konuk, and Kuchaiev]{renduchintala2023tied}
Adithya Renduchintala, Tugrul Konuk, and Oleksii Kuchaiev.
\newblock Tied-lora: Enhacing parameter efficiency of lora with weight tying.
\newblock \emph{CoRR}, abs/2311.09578, 2023.

\bibitem[Shibata et~al.(1999)Shibata, Kida, Fukamachi, Takeda, Shinohara, Shinohara, and Arikawa]{shibata1999byte}
Yusuxke Shibata, Takuya Kida, Shuichi Fukamachi, Masayuki Takeda, Ayumi Shinohara, Takeshi Shinohara, and Setsuo Arikawa.
\newblock Byte pair encoding: A text compression scheme that accelerates pattern matching.
\newblock 1999.

\bibitem[Smith et~al.(2021)Smith, Dherin, Barrett, and De]{smith2021origin}
Samuel~L Smith, Benoit Dherin, David~GT Barrett, and Soham De.
\newblock On the origin of implicit regularization in stochastic gradient descent.
\newblock \emph{arXiv preprint arXiv:2101.12176}, 2021.

\bibitem[Srivastava et~al.(2014)Srivastava, Hinton, Krizhevsky, Sutskever, and Salakhutdinov]{srivastava2014dropout}
Nitish Srivastava, Geoffrey Hinton, Alex Krizhevsky, Ilya Sutskever, and Ruslan Salakhutdinov.
\newblock Dropout: A simple way to prevent neural networks from overfitting.
\newblock \emph{Journal of machine learning research}, 15\penalty0 (1):\penalty0 1929--1958, 2014.

\bibitem[Team et~al.(2024)Team, Mesnard, Hardin, Dadashi, Bhupatiraju, Pathak, Sifre, Rivi{\`e}re, Kale, Love, et~al.]{team2024gemma}
Gemma Team, Thomas Mesnard, Cassidy Hardin, Robert Dadashi, Surya Bhupatiraju, Shreya Pathak, Laurent Sifre, Morgane Rivi{\`e}re, Mihir~Sanjay Kale, Juliette Love, et~al.
\newblock Gemma: Open models based on gemini research and technology.
\newblock \emph{arXiv preprint arXiv:2403.08295}, 2024.

\bibitem[Touvron et~al.(2023)Touvron, Martin, Stone, Albert, Almahairi, Babaei, Bashlykov, Batra, Bhargava, Bhosale, et~al.]{touvron2023llama}
Hugo Touvron, Louis Martin, Kevin Stone, Peter Albert, Amjad Almahairi, Yasmine Babaei, Nikolay Bashlykov, Soumya Batra, Prajjwal Bhargava, Shruti Bhosale, et~al.
\newblock Llama 2: Open foundation and fine-tuned chat models.
\newblock \emph{arXiv preprint arXiv:2307.09288}, 2023.

\bibitem[Wager et~al.(2013)Wager, Wang, and Liang]{wager2013dropout}
Stefan Wager, Sida Wang, and Percy Liang.
\newblock Dropout training as adaptive regularization.
\newblock In Christopher J.~C. Burges, Léon Bottou, Zoubin Ghahramani, and Kilian~Q. Weinberger (eds.), \emph{NIPS}, pp.\  351--359, 2013.

\bibitem[Wang \& Manning(2013)Wang and Manning]{wang2013fast}
Sida Wang and Christopher Manning.
\newblock Fast dropout training.
\newblock In \emph{Proceedings of the 30th International Conference on Machine Learning}, pp.\  118--126, 2013.

\bibitem[Wang et~al.(2023)Wang, Wu, Dabral, Zhang, Brown, Lu, Liu, Liang, Pang, Bendersky, et~al.]{wang2023non}
Yaqing Wang, Jialin Wu, Tanmaya Dabral, Jiageng Zhang, Geoff Brown, Chun-Ta Lu, Frederick Liu, Yi~Liang, Bo~Pang, Michael Bendersky, et~al.
\newblock Non-intrusive adaptation: Input-centric parameter-efficient fine-tuning for versatile multimodal modeling.
\newblock \emph{arXiv preprint arXiv:2310.12100}, 2023.

\bibitem[Yeh et~al.(2024)Yeh, Hsieh, Gao, Yang, Oh, and Gong]{yeh2024text}
Shih-Ying Yeh, Yu-Guan Hsieh, Zhidong Gao, Bernard B.~W. Yang, Giyeong Oh, and Yanmin Gong.
\newblock Navigating text-to-image customization: From lycoris fine-tuning to model evaluation.
\newblock In \emph{ICLR} \citet{yeh2024text}.

\bibitem[Zhang \& Pilanci(2024)Zhang and Pilanci]{zhang2024riemannian}
Fangzhao Zhang and Mert Pilanci.
\newblock Riemannian preconditioned lora for fine-tuning foundation models.
\newblock \emph{arXiv preprint arXiv:2402.02347}, 2024.

\bibitem[Zhang et~al.(2023)Zhang, Chen, Bukharin, He, Cheng, Chen, and Zhao]{zhang2023adaptive}
Qingru Zhang, Minshuo Chen, Alexander Bukharin, Pengcheng He, Yu~Cheng, Weizhu Chen, and Tuo Zhao.
\newblock Adaptive budget allocation for parameter-efficient fine-tuning.
\newblock In \emph{ICLR} \citet{zhang2023adaptive}.

\bibitem[Zhang et~al.(2010)Zhang, Jin, and Zhou]{zhang2010understanding}
Yin Zhang, Rong Jin, and Zhi-Hua Zhou.
\newblock Understanding bag-of-words model: a statistical framework.
\newblock \emph{International journal of machine learning and cybernetics}, 1:\penalty0 43--52, 2010.

\bibitem[Zhao et~al.(2024)Zhao, Zhang, Chen, Wang, Anandkumar, and Tian]{zhao2024galore}
Jiawei Zhao, Zhenyu Zhang, Beidi Chen, Zhangyang Wang, Anima Anandkumar, and Yuandong Tian.
\newblock Galore: Memory-efficient llm training by gradient low-rank projection.
\newblock \emph{arXiv preprint arXiv:2403.03507}, 2024.

\end{thebibliography}
\bibliographystyle{iclr2025_conference}

\newpage

\appendix

\section{Proof of upper bound}
\label{proof:upper_bound}

\begin{proof}
    To simplify the derivations, we will denote $\mY - \mX\mW - ((\mX\mA)\odot \mV_n)\mB$ by $\mZ_n$.
\begin{align*}
    &\mathbb{E}\left[ \left| \frac{1}{N}\sum_{n=1}^{N}\| \mZ_n\|_F^2-\mathbb{E}\left[\| \mZ\|_F^2\right]\right|\right]\\
    =&\mathbb{E}\left[\sqrt{ \left( \frac{1}{N}\sum_{n=1}^{N}\| \mZ_n\|_F^2-\mathbb{E}\left[\| \mZ\|_F^2\right]\right)^2}\right]\\
    \leq & \sqrt{\mathbb{E}\left[ \left( \frac{1}{N}\sum_{n=1}^{N}\| \mZ_n\|_F^2-\mathbb{E}\left[\| \mZ\|_F^2\right]\right)^2\right]}\\
    =& \sqrt{\mathbb{E}\left[ \left( \frac{1}{N}\sum_{n=1}^{N}\| \mZ_n\|_F^2\right)^2\right]-\mathbb{E}\left[\| \mZ\|_F^2\right]^2}\\
    % =& \sqrt{\mathbb{E}\left[ \left( \frac{1}{N}\sum_{n=1}^{N}\| \mZ_n\|_F^2\right)^2\right]-\mathbb{E}\left[\| \mZ\|_F^2\right]^2}\\
    =& \sqrt{\frac{1}{N^2}\sum_{n=1}^{N}\mathbb{E}\left[\| \mZ_n\|_F^4\right]-\frac{1}{N}\mathbb{E}\left[\| \mZ\|_F^2\right]^2}\\
    =& \sqrt{\frac{1}{N^2}\sum_{n=1}^{N}\left(\mathbb{E}\left[\| \mZ_n\|_F^4\right]-\mathbb{E}\left[\| \mZ\|_F^2\right]^2\right)}\\
    =& \sqrt{\frac{1}{N}Var\left[\| \mZ\|_F^2\right]}\\
    =& \frac{Std\left[\| \mZ\|_F^2\right]}{\sqrt{N}}
\end{align*}
\end{proof}

\section{Finetuning Accuracy At Various Dropout Rates}
\Cref{tab:accuracy-epoch-dropout} contains accuracy of various dropout rates at different number of epochs.

\begin{table}
  \begin{tabular}{cccccc}
    \toprule
dropout & 0.0 & 0.05 & 0.1 & 0.2 & 0.4 \\
    \midrule
Acc. at epoch=3 & 77.92 & 77.96 & 77.68 & 77.44 & 77.22 \\
Acc. at epoch=10 & 78.00 & 79.95 & 80.50 & 80.77 & 80.65 \\
Max acc. & 79.81 & 80.27 & 80.54 & 80.78 & 80.65 \\
Epoch of max acc. & 6 & 8 & 8 & 9 & 10 \\
    \bottomrule
    \end{tabular}
  \caption{\small Reprise of \cref{fig:accuracy-epoch-dropout} Left depicting that the epoch at which the LoRA fine-tuned model reaches the best accuracy increases with the Dropout rate, i.e., the larger the probability to drop dimensions, the more regularization is applied and the better the final performance--but only after very long fine-tuning.}%
  \label{tab:accuracy-epoch-dropout}
\end{table}

\section{Adaptive Function}
\begin{figure}[t]
    \centering
    \includegraphics[width=0.35\textwidth]{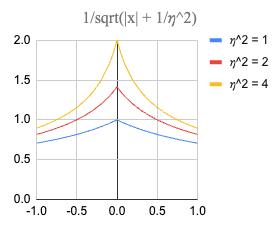} \\
    \caption{Adaptive function $1/\sqrt{|x| + 1/\eta^2}$ for $\eta^2=1, 2, 4$.}
    \label{fig:adapt-lr}
\end{figure}

\Cref{fig:adapt-lr} is an adaptive function that provides output value when $|x|=0$, and then tapers down when $|x| > 0$.

\section{Perception Tasks}
\Cref{tab:nlu_full} shows the accuracy data of all of our experiments on perception tasks. Each cell is an average of 5 runs. ALLoRA is universally better than LoRA in terms of accuracy. 

\begin{table}
\setlength{\tabcolsep}{3pt}
    \centering
    \caption{Accuracy comparison of various models, datasets, and learning rates between ALLoRA and plain LoRA.
    ALLoRA+D is the version of ALLoRA with $0.05$ dropout rate. LoRA's learning rate is $\eta \times 1e-4$, ALLoRA's and 
    ALLoRA+D's base learning rate is $1e-4$. Each cell is an average over 5 runs.}
    \label{tab:nlu_full}
    \small
\begin{tabular}{ccccccccccc}
\toprule
\multirow[m]{2}{*}{Learning Rate} & \multirow[m]{2}{*}{Method}
                 & \multicolumn{3}{c}{Qwen2-0.5B} & \multicolumn{3}{c}{Snowflake-Artic-L} & \multicolumn{3}{c}{OpenELM-450M} \\
               & & b-in-b & emotion & rotten & emotion & rotten & yelp & emotion & rotten & yelp\\
\midrule
\multirow[m]{3}{*}{\begin{tabular}{c}$\eta^2 = 1.0$ \\ $l_b = 1e-4$ \end{tabular}}
               & LoRA            & 70.81 & 35.59 & 53.19 & 87.06 & 77.71 & 63.61 & 84.98 & 87.95 & 71.41 \\
               & ALLoRA+D (ours)   & 71.29 & 37.18 & 53.64 & 86.93 & \textbf{78.29} & 63.55 & \textbf{86.07} & \textbf{88.20} & 71.32 \\
               & ALLoRA (ours) & \textbf{71.53} & \textbf{37.35} & \textbf{54.33} & 86.98 & 78.19 & 63.52 & 86.04 & 88.11 & 71.36 \\
\midrule
\multirow[m]{3}{*}{\begin{tabular}{c}$\eta^2 = 2.0$ \\ $l_b = 1e-4$ \end{tabular}}
               & LoRA            & 73.70 & 37.76 & 54.32 & 88.30 & 79.34 & 64.26 & 90.01 & 88.74 & 71.62 \\
               & ALLoRA+D (ours)   & 73.77 & \textbf{38.52} & 54.67 & 88.23 & \textbf{79.44} & 64.24 & \textbf{90.01} & 88.95 & 71.63 \\
               & ALLoRA (ours) & \textbf{73.99} & 38.09 & \textbf{55.37} & \textbf{88.52} & 79.16 & \textbf{64.33} & 89.93 & \textbf{89.04} & \textbf{71.68} \\
\midrule
\multirow[m]{3}{*}{\begin{tabular}{c}$\eta^2 = 4.0$ \\ $l_b = 1e-4$ \end{tabular}}
               & LoRA            & 75.60 & 38.68 & 54.90 & 88.95 & 80.04 & 64.76 & 91.33 & 89.55 & 71.79 \\
               & ALLoRA+D (ours)   & \textbf{75.72} & 39.00 & \textbf{55.83} & \textbf{89.08} & \textbf{80.24} & 64.61 & 91.27 & \textbf{89.61} & \textbf{71.83} \\
               & ALLoRA (ours) & 75.52 & \textbf{39.47} & 55.68 & 88.91 & \textbf{80.24} & \textbf{64.76} & \textbf{91.43} & 89.47 & 71.82 \\
\bottomrule
\end{tabular}
\end{table}

\begin{table}
\setlength{\tabcolsep}{3pt}
    \centering
    \caption{Accuracy comparison of various models, datasets, and learning rates between ALLoRA and other adaptive approaches.
    for adaptive learning rate approaches, i.e., ALLoRA and ALLoRA-OD, base learning rate is $1e-4$. For adaptive scaling factor,
    i.e., ASF-LoRA, learning rate is fixed at $1e-4$, an adaptive scaling factor $1/\sqrt{|x| + 1/\eta^2}$ is applied.
    For LoRA, a fixed scaling factor $\eta$ is applied, and learning rate is fixed at $1e-4$. Each cell is an average over 5 runs.}
    \label{tab:output-dep}
    \small
\begin{tabular}{cccccccc}
\toprule
\multirow[m]{2}{*}{$\eta^2$} & \multirow[m]{2}{*}{Method}
                 & \multicolumn{2}{c}{Qwen2-0.5B} & \multicolumn{2}{c}{Snowflake-Artic-L} & \multicolumn{2}{c}{OpenELM-450M} \\
               & & emotion & rotten & emotion & rotten & emotion & rotten \\
\midrule
\multirow[m]{4}{*}{$1.0$}
               & ALLoRA (ours)      & \textbf{37.35} & \textbf{54.33} & 86.98 & \textbf{78.19} & 86.04 & \textbf{88.11} \\
               & ALLoRA-OD          & 36.91 & 52.31 & 87.24 & 77.47 & 86.31 & 88.11 \\
               & ASF-LoRA           & 36.40 & 53.43 & 87.30 & 78.03 & 83.44 & 87.58 \\
               & LoRA $\times \eta$ & 35.59 & 53.19 & 87.06 & 77.71 & 84.98 & 87.95 \\
\midrule
\multirow[m]{4}{*}{$2.0$}
               & ALLoRA (ours)      & 38.09 & \textbf{55.37} & \textbf{88.52} & \textbf{79.16} & \textbf{89.93} & \textbf{89.04} \\
               & ALLoRA-OD          & 38.15 & 53.55 & 88.22 & 79.16 & 89.86 & 89.02 \\
               & ASF-LoRA           & 37.20 & 54.17 & 87.74 & 78.18 & 86.26 & 88.22 \\
               & LoRA $\times \eta$ & 36.61 & 54.47 & 87.62 & 78.72 & 88.77 & 88.39 \\
\midrule
\multirow[m]{4}{*}{$4.0$}
               & ALLoRA (ours)      & \textbf{39.47} & \textbf{55.68} & 88.91 & \textbf{80.24} & \textbf{91.43} & \textbf{89.47} \\
               & ALLoRA-OD          & 38.67 & 54.95 & 88.97 & 80.21 & 91.27 & 89.38 \\
               & ASF-LoRA           & 38.45 & 54.41 & 87.97 & 79.23 & 89.64 & 88.80 \\
               & LoRA $\times \eta$ & 37.63 & 53.70 & 88.49 & 79.31 & 90.25 & 89.08 \\
\bottomrule
\end{tabular}
\end{table}

\section{Ablation}
\Cref{tab:output-dep} shows the accuracy data of all of our ablation study on perception tasks. Each cell is an average of 5 runs. ALLoRA is universally better than the rest in the family. 

% \section{Adaptive Learning Examples}\label{sec:ale}

\clearpage

\definecolor{codegreen}{rgb}{0,0.6,0}
\definecolor{codegray}{rgb}{0.5,0.5,0.5}
\definecolor{codepurple}{rgb}{0.58,0,0.82}
\definecolor{codeblue}{rgb}{0,0,0.98}
\definecolor{backcolour}{rgb}{0.98,0.98,0.96}

\lstdefinestyle{pythoncode}{
    backgroundcolor=\color{backcolour},   
    commentstyle=\color{codegreen},
    keywordstyle=\color{magenta},
    numberstyle=\tiny\color{codegray},
    stringstyle=\color{codepurple},
    basicstyle=\ttfamily\footnotesize,
    breakatwhitespace=false,         
    breaklines=true,                 
    captionpos=b,                    
    keepspaces=true,                 
    numbers=left,                    
    numbersep=5pt,                  
    showspaces=false,                
    showstringspaces=false,
    showtabs=false,                  
    tabsize=2
}

\lstset{style=pythoncode}

\section{Code}
\small
\begin{lstlisting}[caption={ALLoRA Code},captionpos=b, label=code:allora,language=python]
class ALLoRA(torch.autograd.Function):
  rsq_scale = 1. / 4.  # 1 / \eta^2

  @staticmethod
  def forward(ctx, input_x, weight_A, weight_B):
    output = input_x @ weight_A.t() @ weight_B.t()
    norms = torch.norm(weight_B @ weight_A, dim=1)
    ctx.save_for_backward(input_x, weight_A, weight_B, norms)
    return output

  @staticmethod
  def backward(ctx, grad_output):
    input_x, weight_A, weight_B, norms = ctx.saved_tensors
    accelerate = 1. / torch.sqrt(norms + LinearLayer2.rsq_scale)
    grad_input = grad_output @ weight_B @ weight_A  
    temp = grad_output.mul(accelerate) @ weight_B    
    temp = torch.transpose(temp, 1, 2)
    grad_weight_A = temp @ input_x
    temp = grad_output.mul(accelerate).transpose(1, 2)  
    grad_weight_B = temp @ (input_x @ weight_A.t())
    return grad_input, grad_weight_A, grad_weight_B
\end{lstlisting}

\end{document}